\documentclass[10pt,twocolumn,letterpaper]{article}

\usepackage{iccv}
\usepackage{times}
\usepackage{epsfig}
\usepackage{graphicx}
\usepackage{amsmath}
\usepackage{amssymb}
\usepackage{booktabs}
\usepackage{arydshln}
\usepackage{threeparttable}
\usepackage{bm}
\usepackage{hyperref}
\hypersetup{
    colorlinks=true,
    linkcolor=blue,
    filecolor=magenta,      
    urlcolor=magenta,
}
 
\iccvfinalcopy 

\def\iccvPaperID{732} 

\ificcvfinal\fi
\begin{document}

\title{Cross View Fusion for $3$D Human Pose Estimation}

\author{Haibo Qiu \thanks{This work is done when Haibo Qiu is an intern at Microsoft Research Asia.}\\
University of Science and Technology of China\\
{\tt\small haibo-qiu@outlook.com}
\and
Chunyu Wang\\
Microsoft Research Asia\\
{\tt\small chnuwa@microsoft.com}
\and
Jingdong Wang\\
Microsoft Research Asia\\
{\tt\small jingdw@microsoft.com}
\and
Naiyan Wang\\
TuSimple\\
{\tt\small winsty@gmail.com}
\and
Wenjun Zeng\\
Microsoft Research Asia\\
{\tt\small wezeng@microsoft.com}
}

\maketitle

\begin{abstract}
We present an approach to recover absolute $3$D human poses from multi-view images by incorporating multi-view geometric priors in our model. It consists of two separate steps: (1) estimating the $2$D poses in multi-view images and (2) recovering the $3$D poses from the multi-view $2$D poses.
First,
we introduce a cross-view fusion scheme into CNN to jointly estimate $2$D poses for multiple views.
Consequently, the $2$D pose estimation for each view already benefits from other views.
Second,
we present a recursive Pictorial Structure Model to recover the $3$D pose from the multi-view $2$D poses.
It gradually improves the accuracy of $3$D pose with affordable computational cost.
We test our method on two public datasets H36M and Total Capture. The Mean Per Joint Position Errors on the two datasets are $26$mm and $29$mm, which outperforms the state-of-the-arts remarkably ($26$mm vs $52$mm, $29$mm vs $35$mm). Our code is released at \url{https://github.com/microsoft/multiview-human-pose-estimation-pytorch}.
\end{abstract}

\section{Introduction}
The task of $3$D pose estimation has made significant progress due to the introduction of deep neural networks. Most efforts \cite{martinez2017simple, kanazawa2018end,zhou2017towards,pavlakos2017coarse,tome2017lifting,Rhodin_2018_ECCV,wang2014robust,wang2018robust,lcn2019} have been devoted to estimating \textit{relative} $3$D poses from monocular images. The estimated poses are centered around the pelvis joint thus do not know their absolute locations in the environment (world coordinate system).

In this paper, we tackle the problem of estimating absolute $3$D poses in the world coordinate system from multiple cameras \cite{amin2013multi,liu2011markerless,burenius20133D,PavlakosZDD17,belagiannis20143D,rhodin2018learning}. Most works follow the pipeline of first estimating $2$D poses and then recovering $3$D pose from them. However, the latter step usually depends on the performance of the first step which unfortunately often has large errors in practice especially when occlusion or motion blur occurs in images. This poses a big challenge for the final $3$D estimation. 

On the other hand, using the Pictorial Structure Model (PSM) \cite{kostrikov2014depth,PavlakosZDD17,belagiannis20143D} for $3$D pose estimation can alleviate the influence of inaccurate $2$D joints by considering their spatial dependence. It discretizes the space around the root joint by an $N \times N \times N$ grid and assigns each joint to one of the $N^3$ bins (hypotheses). It jointly minimizes the projection error between the estimated $3$D pose and the $2$D pose, along with the discrepancy of the spatial configuration of joints and its prior structures. However, the space discretization causes large quantization errors. For example, when the space surrounding the human is of size $2000$mm and $N$ is $32$, the quantization error is as large as $30$mm. We could reduce the error by increasing $N$, but the inference cost also increases at $O(N^6)$, which is usually intractable.

\begin{figure}
	\centering
	\includegraphics[width=3in]{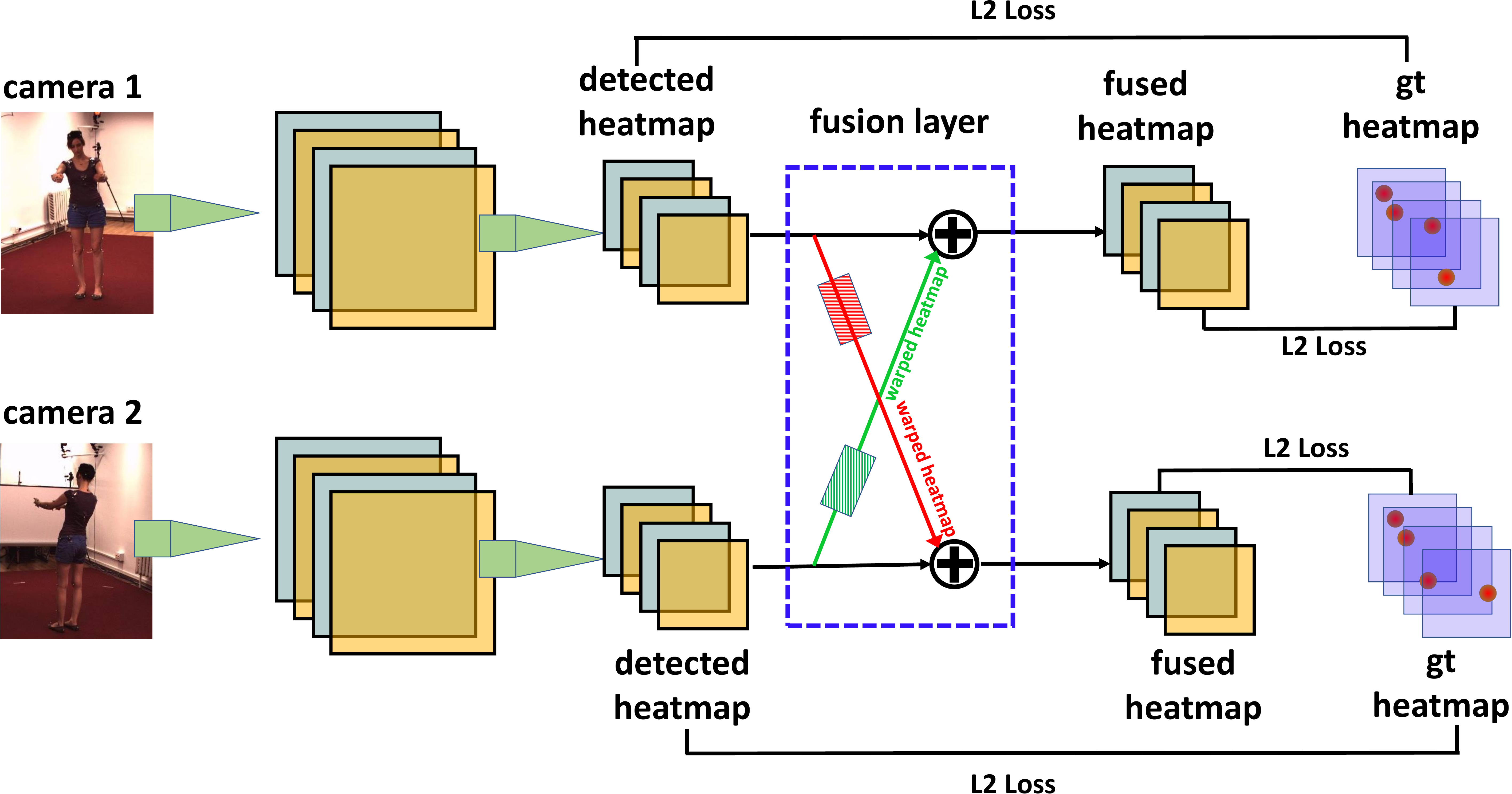}
	\caption{Cross-view fusion for $2$D pose estimation. The images are first fed into a CNN to get initial heatmaps. Then the heatmap of each view is fused with the heatmaps from other views through a fusion layer.  The whole network is learned end-to-end.}
	\label{fig:pipeline}
\end{figure}

Our work aims to address the above challenges. First, we obtain more accurate $2$D poses by jointly estimating them from multiple views using a CNN based approach. It elegantly addresses the challenge of finding the corresponding locations between different views for $2$D pose heatmap fusion. We implement this idea by a fusion neural network as shown in Figure \ref{fig:pipeline}. The fusion network can be integrated with any CNN based $2$D pose estimators in an end-to-end manner without intermediate supervision.

Second, we present Recursive Pictorial Structure Model (RPSM), to recover the $3$D pose from the estimated multi-view $2$D pose heatmaps. Different from PSM which directly discretizes the space into a large number of bins in order to control the quantization error, RPSM \textit{recursively} discretizes the space around each joint location (estimated in the previous iteration) into a \textbf{finer-grained} grid using a \textit{small} number of bins. As a result, the estimated $3$D pose is refined step by step.  Since $N$ in each step is usually small, the inference speed is very fast for a single iteration. In our experiments, RPSM decreases the error by at least $50\%$ compared to PSM with little increase of inference time.

For $2$D pose estimation on the H36M dataset \cite{ionescu2014human3}, the average detection rate over all joints improves from $89\%$ to $96\%$. The improvement is significant for the most challenging ``wrist'' joint. For $3$D pose estimation, changing PSM to RPSM dramatically reduces the average error from $77$mm to $26$mm. Even compared with the state-of-the-art method with an average error $52$mm, our approach also cuts the error in half. We further evaluate our approach on the Total Capture dataset \cite{trumble2017total} to validate its generalization ability. It still outperforms the state-of-the-art \cite{trumble2018deep}.

\section{Related Work}
We first review the related work on multi-view $3$D pose estimation and discuss how they differ from our work. Then we discuss some techniques on feature fusion. 

\paragraph{Multi-view $3$D Pose Estimation} Many approaches \cite{liu2011markerless,gall2010optimization,burenius20133D,PavlakosZDD17,belagiannis20143D,Rhodin_2018_ECCV,rhodin2018learning} are proposed for multi-view pose estimation. They first define a body model represented as simple primitives, and then optimize the model parameters to align the projections of the body model with the image features. These approaches differ in terms of the used image features and optimization algorithms.

We focus on the Pictorial Structure Model (PSM) which is widely used in object detection \cite{felzenszwalb2005pictorial,fischler1973representation} to model the spatial dependence between the object parts. This technique is also used for $2$D \cite{yang2011articulated,chen2014articulated,amin2013multi} and $3$D \cite{burenius20133D,PavlakosZDD17} pose estimation where the parts are the body joints or limbs. In \cite{amin2013multi}, Amin \etal first estimate the $2$D poses in a multi-view setup with PSM and then obtain the $3$D poses by direct triangulation. Later Burenius \etal \cite{burenius20133D} and Pavlakos \etal \cite{PavlakosZDD17} extend PSM to multi-view $3$D human pose estimation. For example, in \cite{PavlakosZDD17}, they first estimate $2$D poses independently for each view and then recover the $3$D pose using PSM. Our work differs from \cite{PavlakosZDD17} in that we extend PSM to a recursive version, \ie RPSM, which efficiently refines the 3D pose estimations step by step. In addition, they \cite{PavlakosZDD17} do not perform cross-view feature fusion as we do.

\paragraph{Multi-image Feature Fusion} Fusing features from different sources is a common practice in the computer vision literature. For example, in \cite{zhu2017flow}, Zhu \etal propose to warp the features of the neighboring frames (in a video sequence) to the current frame according to optical flow in order to robustly detect the objects. Ding \etal \cite{Ding_2018_CVPR} propose to aggregate the multi-scale features which achieves better segmentation accuracy for both large and small objects.  Amin \etal \cite{amin2013multi} propose to estimate $2$D poses by exploring the geometric relation between multi-view images. It differs from our work in that it does not \emph{fuse} features from other views to obtain better $2$D heatmaps. Instead, they use the multi-view $3$D geometric relation to \emph{select} the joint locations from the  ``imperfect'' heatmaps. In \cite{jafarian2018monet}, multi-view consistency is used as a source of supervision to train the pose estimation network. To the best of our knowledge, there is no previous work which fuses multi-view features so as to obtain better $2$D pose heatmaps because it is a challenging task to find the corresponding features across different views which is one of our key contributions of this work.

\begin{figure}
	\centering
	\includegraphics[width=2.2in]{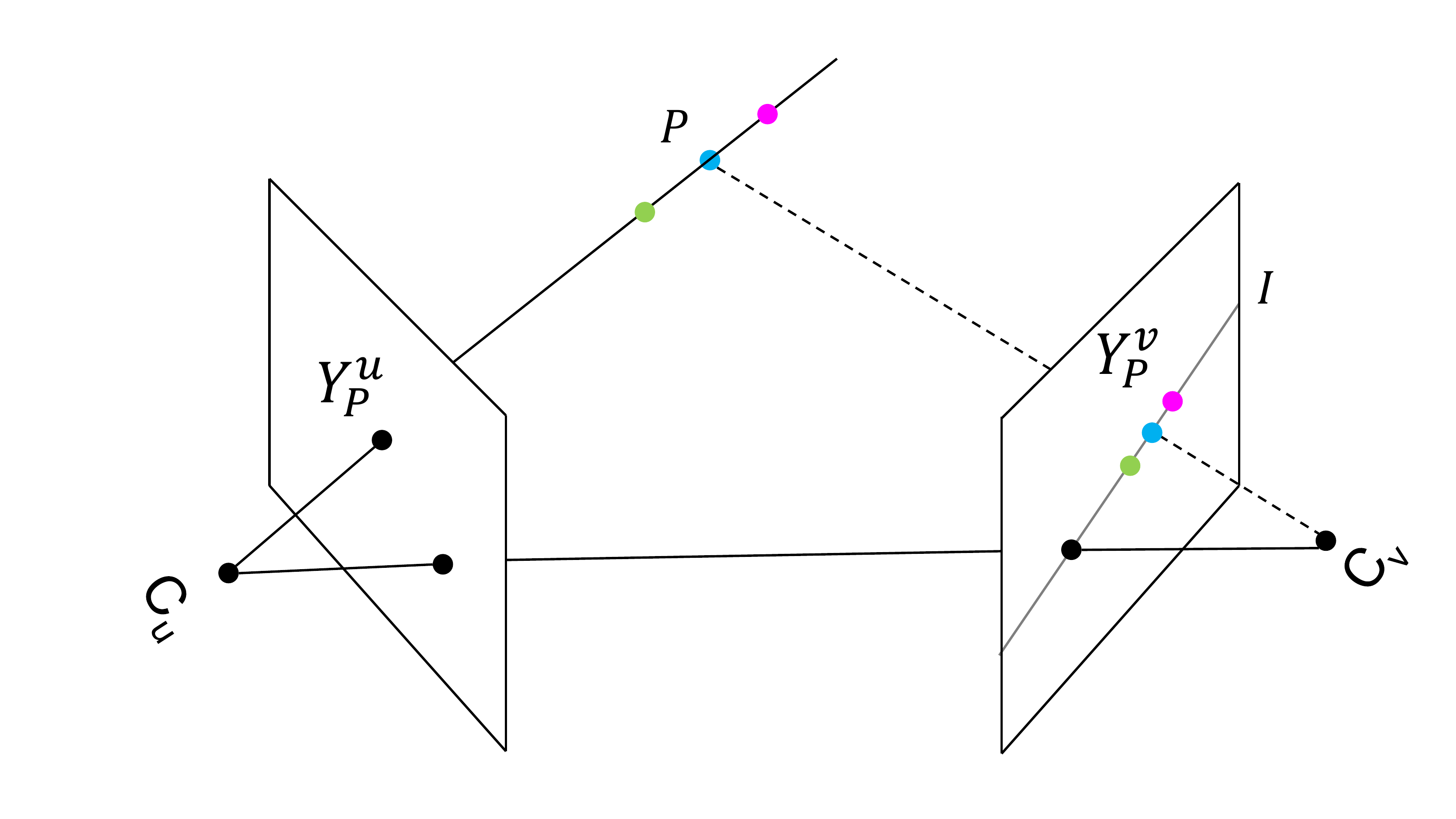}
	\caption{Epipolar geometry: an image point $Y_P^u$ back-projects to a ray in 3D defined by the camera $C_u$ and $Y_P^u$. This line is imaged as $I$ in the camera $C_v$. The 3D point $P$ which projects to $Y_P^u$ must lie on this ray, so the image of $P$ in camera $C_v$ must lie on $I$.}
	\label{fig:epipolar_geometry}
\end{figure}

\section{Cross View Fusion for $2$D Pose Estimation}
Our $2$D pose estimator takes multi-view images as input, generates initial pose heatmaps respectively for each, and then
fuses the heatmaps across different views such that the heatmap of each view benefits from others. The process is accomplished in a single CNN and can be trained end-to-end. Figure \ref{fig:pipeline} shows the pipeline for two-view fusion. Extending it to multi-views is trivial where the heatmap of each view is fused with the heatmaps of all other views. The core of our fusion approach is to find the corresponding features between a pair of views.

Suppose there is a point $\bm{P}$ in $3$D space. See Figure \ref{fig:epipolar_geometry}. Its projections in view $u$ and $v$ are $\bm{Y}_P^u \in \mathcal{Z}^u$ and $\bm{Y}_P^v \in \mathcal{Z}^v$, respectively where $\mathcal{Z}^u$ and $\mathcal{Z}^v$ denote all pixel locations in the two views, respectively. The heatmaps of view $u$ and $v$ are $\mathcal{F}^u=\{\bm{x}_1^u, \cdots, \bm{x}_{|\mathcal{Z}^u|}^u\}$ and $\mathcal{F}^v=\{\bm{x}_1^v, \cdots, \bm{x}_{|\mathcal{Z}^v|}^v\}$. The core idea of fusing a feature in view $u$, say $\bm{x}_i^u$, with the features from $\mathcal{F}^v$ is to establish the correspondence between the two views:
\begin{equation}
    \bm{x}_i^u \leftarrow \bm{x}_i^u + \sum_{j=1}^{|\mathcal{Z}^v|}{\omega_{j,i} \cdot \bm{x}_j^v}, \quad \forall i \in \mathcal{Z}^u,
    \label{eq:update_rule}
\end{equation}
where $\omega_{j,i}$ is a to be determined scalar. Ideally, for a specific $i$, only one $\omega_{j,i}$ should be positive, while the rest are zero. Specifically, $\omega_{j,i}$ is positive when the pixel $i$ in view $u$ and pixel $j$ in view $v$ correspond to the same $3$D point.

Suppose we know only $\bm{Y}_P^u$, how can we find the corresponding point $\bm{Y}_P^v$ in the image of a different view? We know $\bm{Y}_P^v$ is guaranteed to lie on the epipolar line $I$. But since we do not know the depth of $\bm{P}$, which means it may move on the line defined by $\bm{C}_u$ and $\bm{Y}_P^u$, we cannot determine the exact location of $\bm{Y}_P^v$ on $I$. This ambiguity poses a challenge for the cross view fusion.

Our solution is to fuse $\bm{x}_i^u$ with all features on the line $I$. This may sound brutal at the first glance, but is in fact elegant. Since fusion happens in the heatmap layer, ideally, $\bm{x}_j^v$ should have large response at $\bm{Y}_P^v$ (the cyan point) and zeros at other locations on the epipolar line $I$. It means the non-corresponding locations on the line will contribute no or little to the fusion. So fusing all pixels on the epipolar line is a simple yet effective solution.

\begin{figure}
	\centering
	\includegraphics[width=2.4in]{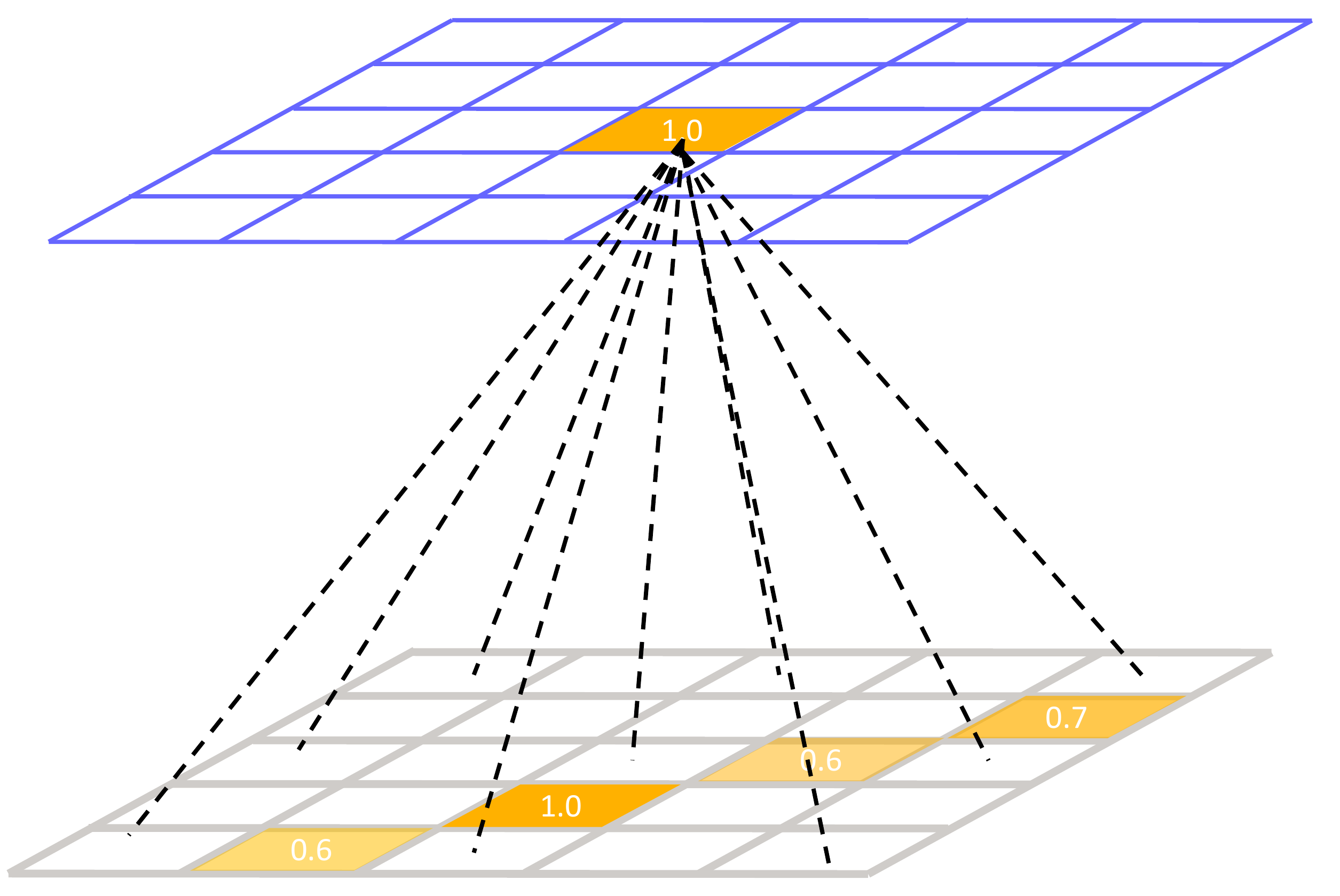}
	\caption{Two-view feature fusion for one channel. The top grid denotes the feature map of view $A$. Each location in view $A$ is connected to all pixels in view $B$ by a weight matrix. The weights are mostly positive for locations on the epipolar line (numbers in the yellow cells). Different locations in view $A$ have different weights because they correspond to different epipolar lines.}
	\label{fig:feature_fusion}
\end{figure}

\subsection{Implementation}
The feature fusion rule (Eq. (\ref{eq:update_rule})) can be interpreted as a fully connected layer imposed on each channel of the pose heatmaps where $\omega$ are the learnable parameters. Figure \ref{fig:feature_fusion} illustrates this idea. 
Different channels of the feature maps, which correspond to different joints, share the same weights because the cross view relations do not depend on the joint types but only depend on the pixel locations in the camera views. Treating feature fusion as a neural network layer enables the end-to-end learning of the weights. 

We investigate two methods to train the network. In the first approach, we clip the positive weights to zero during training if the corresponding locations are off the epipolar line. Negative weights are allowed to represent suppression relations. In the second approach, we allow the network to freely learn the weights from the training data. The final 2D pose estimation results are also similar for the two approaches. So we use the second approach for training because it is simpler.

\subsection{Limitation and Solution}
The learned fusion weights which implicitly encode the information of epipolar geometry are dependent on the camera configurations. As a result, the model trained on a particular camera configuration cannot be directly applied to another different configuration. 

We propose an approach to automatically adapt our model to a new environment without any annotations. 
We adopt a semi-supervised training approach following the previous work \cite{simon2017hand}. First, we train a single view 2D pose estimator \cite{simplebaselines} on the existing datasets such as MPII which have ground truth pose annotations. Then we apply the trained model to the images captured by multiple cameras in the new environment and harvest a set of poses as pseudo labels. Since the estimations may be inaccurate for some images, we propose to use multi-view consistency to filter the incorrect labels. We keep the labels which are consistent across different views following \cite{simon2017hand}. In training the cross view fusion network, we do not enforce supervision on the filtered joints. We will evaluate this approach in the experiment section.

\section{RPSM for Multi-view $3$D Pose Estimation}
We represent a human body as a graphical model with $M$ random variables $\mathcal{J}=\{\bm{J}_1,
\bm{J}_2,\cdots,\bm{J}_M\}$ in which each variable corresponds to a body joint. Each variable $J_i$ defines a state vector $\bm{J}_i=[x_i, y_i, z_i]$ as the $3$D position of the body joint in the world coordinate system and takes its value from a discrete state space. See Figure \ref{fig:body_model}. An edge between two variables denotes their conditional dependence  and can be interpreted as a physical constraint.

\subsection{Pictorial Structure Model}
\begin{figure}
	\centering
	\includegraphics[width=1.8in]{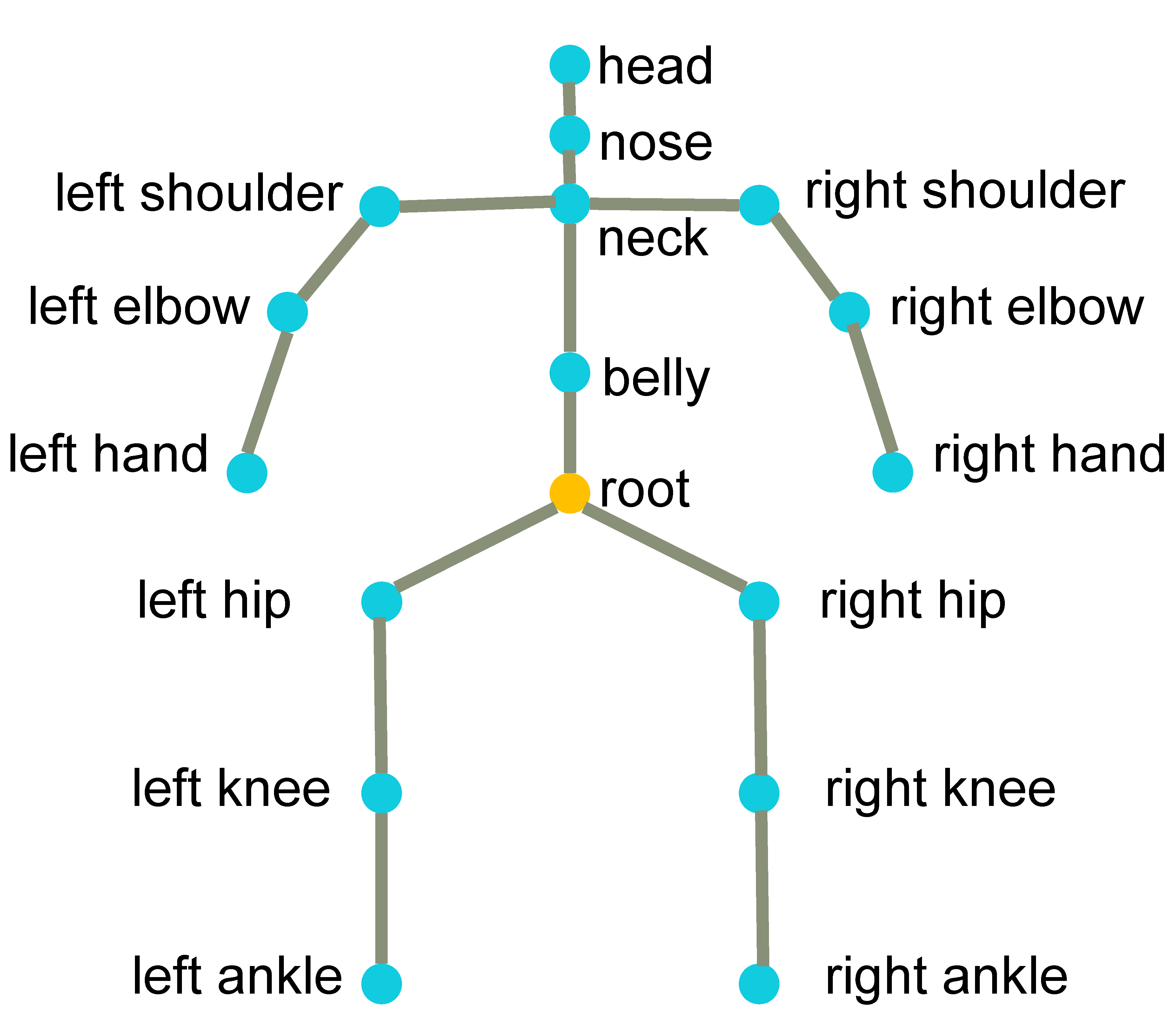}
	\caption{Graphical model of human body used in our experiments. There are $17$ variables and $16$ edges. }
	\label{fig:body_model}
\end{figure}

Given a configuration of $3$D pose $\mathcal{J}$ and multi-view $2$D pose heatmaps $\mathcal{F}$, the posterior becomes \cite{belagiannis20143D}:
\begin{equation}
    p(\mathcal{J} | \mathcal{F})= \frac{1}{Z(\mathcal{F})} \prod_{i=1}^M{\phi_i^{\text{conf}}(\bm{J}_i, \mathcal{F})}\prod_{(m, n) \in \mathcal{E}}{\psi^{\text{limb}}(\bm{J}_m, \bm{J}_n)},
    \label{eq:psm}
\end{equation}
where $Z(\mathcal{F})$ is the partition function and $\mathcal{E}$ are the graph edges as shown in Figure \ref{fig:body_model}. The unary potential functions $\phi_i^{\text{conf}}(\bm{J}_i, \mathcal{F})$ are computed based on the previously estimated multi-view $2$D pose heatmaps $\mathcal{F}$. The pairwise potential functions $\psi^{\text{limb}}(\bm{J}_m, \bm{J}_n)$ encode the limb length constraints between the joints.

\paragraph{Discrete state space} We first triangulate the $3$D location of the root joint using its $2$D locations detected in all views. Then the state space of the $3$D pose is constrained to be within a $3$D bounding volume centered at the root joint. The edge length $s$ of the volume is set to be $2000$mm. The volume is discretized by an $N \times N \times N$ grid $\mathcal{G}$. All body joints share the same state space $\mathcal{G}$ which consists of $N^3$ discrete locations (bins).

\paragraph{Unary potentials} Every body joint hypothesis, \ie a bin in the grid $\mathcal{G}$, is defined by its $3$D position in the world coordinate system. We project it to the pixel coordinate system of all camera views using the camera parameters, and get the corresponding joint confidence from $\mathcal{F}$. We compute the average confidence over all camera views as the unary potential for the hypothesis.

\paragraph{Pairwise potentials}
Offline, for each pair of joints ($\bm{J}_m$,$\bm{J}_n$) in the edge set $\mathcal{E}$, we compute the average distance $\Tilde{l_{m,n}}$ on the training set as limb length priors. During inference, the pairwise potential is defined as:
\begin{equation}
    \psi^{\text{limb}}(\mathbf{J}_m, \mathbf{J}_n) = 
    \left\{
                \begin{array}{ll}
                  1, \quad \text{if} \quad l_{m,n} \in [\Tilde{l_{m,n}}-\epsilon, \Tilde{l_{m,n}}+\epsilon] \\
                  0, \quad \text{otherwise}
                \end{array}
              \right.,
\end{equation}
where $l_{m,n}$ is the distance between $\bm{J}_m$ and $\bm{J}_n$.
The pairwise term favors $3$D poses having reasonable limb lengths. In our experiments, $\epsilon$ is set to be $150$mm.

\paragraph{Inference}
The final step is to maximize the posterior (Eq. (\ref{eq:psm})) over the discrete state space. Because the graph is acyclic, it can be optimized by dynamic programming with global optimum guarantee. The computational complexity is of the order of $\mathcal{O}{(N^6)}$.

\subsection{Recursive Pictorial Structure Model}
The PSM model suffers from large quantization errors caused by space discretization. For example, when we set $N=32$ as in the previous work, the quantization error is as large as $30$mm (\ie $\frac{s}{32 \times 2}$ where $s=2000$ is the edge length of the bounding volume). Increasing $N$ can reduce the quantization error, but the computation time quickly becomes intractable. For example, if $N=64$, the inference speed will be $64=(\frac{64}{32})^6$ times slower.

Instead of using a large $N$ in one iteration, we propose to recursively refine the joint locations through a multiple stage process and use a small $N$ in each stage. In the first stage ($t=0$), we discretize the $3$D bounding volume space around the triangulated root joint using a coarse grid ($N=16$) and obtain an initial $3$D pose estimation $L=(L_1,\cdots,L_M)$ using the PSM approach.

\begin{figure}
	\centering
	\includegraphics[width=3in]{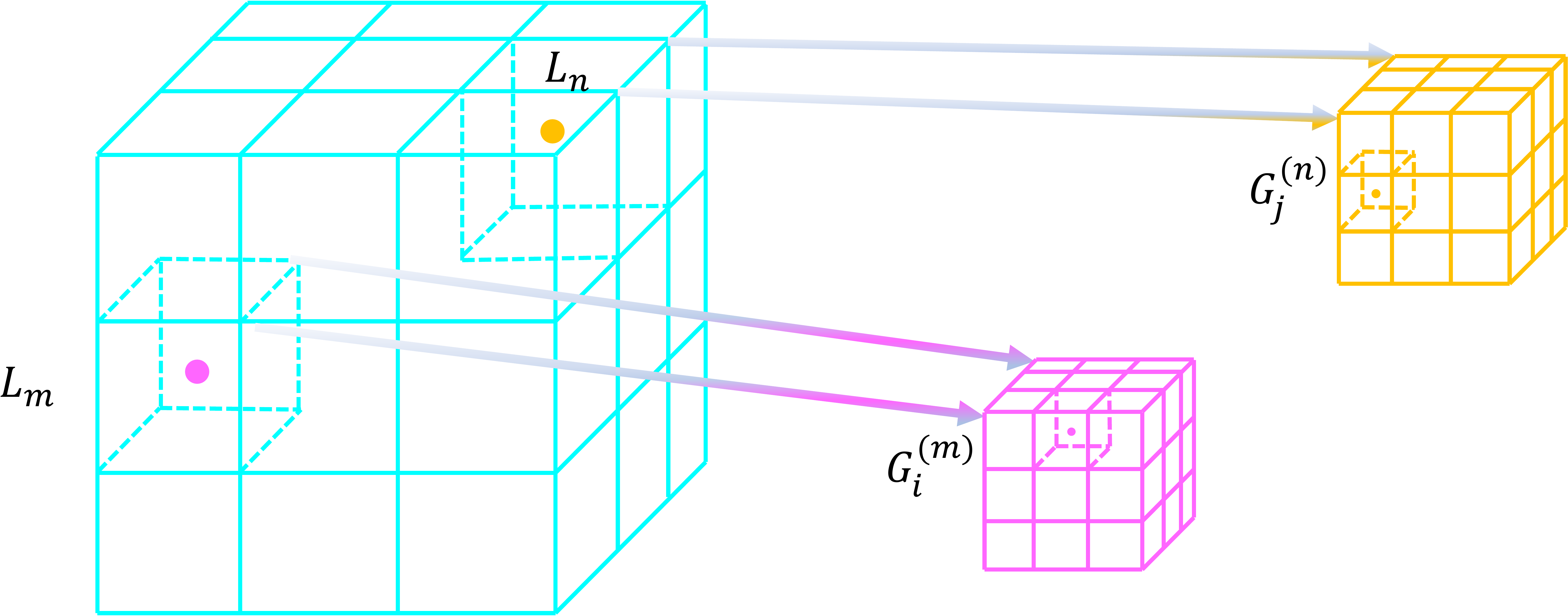}
	\caption{Illustration of the recursive pictorial structure model. Suppose we have estimated the coarse locations $L_m$ and $L_n$ for the two joints $J_m$ and $J_n$, respectively, in the previous iteration. Then we divide the space around the two joints into finer-grained grids and estimate more precise locations.}
	\label{fig:recursive}
\end{figure}

Fo the following stages ($t \ge 1$), for each joint $J_i$, we discretize the space around its current location $L_i$ into an $2 \times 2 \times 2$ grid $G^{(i)}$. The space discretization here differs from PSM in two-fold. First, different joints have their own grids but in PSM all joints share the same grid. See Figure \ref{fig:recursive} for illustration of the idea. Second, the edge length of the bounding volume decreases with iterations: $s_{t}=\frac{s_{t-1}}{N}$. That is the main reason why the grid becomes finer-grained compared to the previous stage.

Instead of refining each joint independently, we simultaneously refine all joints considering their spatial relations. Recall that we know the center locations, sizes and the number of bins of the grids. So we can calculate the location of every bin in the grids with which we can compute the unary and pairwise potentials. It is worth noting that the pairwise potentials should be computed on the fly because it depends on the previously estimated locations. However, because we set $N$ to be a small number (two in our experiments), this computation is fast. 

\subsection{Relation to Bundle Adjustment \cite{triggs1999bundle}}
Bundle adjustment \cite{triggs1999bundle} is also a popular tool for refining $3$D reconstructions. RPSM differs from it in two aspects. First, they reach different local optimums due to their unique ways of space exploration. Bundle adjustment explores in an incremental way while RPSM explores in a divide and conquer way. Second, computing gradients by finite-difference in bundle adjustment is not stable because most entries of heatmaps are zeros.

\section{Datasets and Metrics}
\paragraph{The H36M Dataset \cite{ionescu2014human3}} We use a cross-subject evaluation scheme where subjects $1,5,6,7,8$ are used for training and $9,11$ for testing. We train a single fusion model for all subjects because their camera parameters are similar. In some experiments (which will be clearly stated), we also use the MPII dataset \cite{andriluka20142D} to augment the training data. Since this dataset only has monocular images, we do not train the fusion layer on these images. 

\paragraph{The Total Capture Dataset \cite{trumble2017total}} we also evaluate our approach on the Total Capture dataset to validate its general applicability to other datasets. Following the previous work \cite{trumble2017total}, the training set consists of ``ROM1,2,3", ``Walking1,3", ``Freestyle1,2", ``Acting1,2", ``Running1" on subjects 1,2 and 3. The testing set consists of ``Freestyle3 (\textbf{FS3})", ``Acting3 (\textbf{A3})" and ``Walking2 (\textbf{W2})" on subjects 1,2,3,4 and 5. We use the data of four cameras (1,3,5,7) in experiments.  We do not use the IMU sensors. We do not use the MPII dataset for training in this experiment. The hyper-parameters for training the network are kept the same as those on the H36M dataset.

\paragraph{Metrics} The $2$D pose estimation accuracy is measured by Joint Detection Rate (JDR). If the distance between the estimated and the groundtruth locations is smaller than a threshold, we regard this joint as successfully detected. The threshold is set to be half of the head size as in \cite{andriluka20142D}. JDR is the percentage of the successfully detected joints.

The $3$D pose estimation accuracy is measured by Mean Per Joint Position Error (MPJPE) between the groundtruth $3$D pose $y=[p^3_1,\cdots,p^3_M]$ and the estimated $3$D pose $\bar{y}=[\bar{p^3_1},\cdots,\bar{p^3_M}]$: $\text{MPJPE}=\frac{1}{M} \sum_{i=1}^M \|p^3_i-\bar{p^3_i}\|_2$
We do not align the estimated $3$D poses to the ground truth. This is referred to as protocol 1 in \cite{martinez2017simple,tome2018rethinking}

\begin{table}[]
\center
\caption{This table shows the \textbf{$2$D} pose estimation accuracy on the H36M dataset. ``+MPII" means we train on ``H36M+MPII". We show JDR (\%) for six important joints due to space limitation.}
\label{table:pose$2$D}
{\scriptsize{
\begin{tabular}{l c c c c c c c c}
\hline
Method &  \begin{tabular}{@{}c@{}}Training \\ Dataset \end{tabular}  & Shlder & Elb & Wri & Hip & Knee & Ankle \\
\hline
\textit{Single} & H36M & 88.50 & 88.94 & 85.72 & 90.37 & 94.04 & 90.11\\
\textit{Sum} & H36M & 91.36 & 91.23 & 89.63 & 96.19 & 94.14 & 90.38\\
\textit{Max} & H36M & 92.67 & 92.45 & 91.57 & 97.69 & 95.01 & 91.88\\
\textit{Ours} & H36M & \textbf{95.58} & \textbf{95.83} & \textbf{95.01} & \textbf{99.36} & \textbf{97.96} & \textbf{94.75}\\
\hline
\textit{Single} & +MPII & 97.38 & 93.54 & 89.33 & 99.01 & 95.10 & 91.96\\
\textit{Ours} & +MPII & \textbf{98.97} & \textbf{98.10} & \textbf{97.20} & \textbf{99.85} & \textbf{98.87} & \textbf{95.11}\\
\hline
\end{tabular}
}}
\end{table}

\begin{table}[]
\center
\caption{This table shows the \textbf{$3$D} pose estimation error MPJPE ($mm$) on H36M when different datasets are used for training. ``+MPII" means we use a combined dataset ``H36M+MPII" for training. $3$D poses are obtained by direct triangulation.}
\label{table:pose$3$D}
{\scriptsize{
\begin{tabular}{l c c c c c c c c}
\hline
Method & \begin{tabular}{@{}c@{}}Training \\ Dataset \end{tabular} & Shlder & Elb & Wri & Hip & Knee & Ankle \\
\hline
\textit{Single} & H36M & 59.70 & 89.56 & 313.25 & 69.35 & 76.34 & 120.97\\
\textit{Ours} & H36M & \textbf{42.97} & \textbf{49.83} & \textbf{70.65} & \textbf{24.28} & \textbf{34.42} & \textbf{52.13}\\
\hline
\textit{Single} & +MPII & 30.82 & 38.32 & 64.18 & 24.70 & 38.38 & 62.92\\
\textit{Ours} & +MPII & \textbf{28.99} & \textbf{29.96} & \textbf{34.28} & \textbf{20.65} & \textbf{29.71} & \textbf{47.73}\\
\hline
\end{tabular}
}}
\end{table}

\section{Experiments on $2$D Pose Estimation}
\subsection{Implementation Details}
We adopt the network proposed in \cite{simplebaselines} as our base network and use ResNet-152 as its backbone, which was pre-trained on the ImageNet classification dataset. The input image size is $320 \times 320$ and the  resolution of the heatmap is $80 \times 80$. We use heatmaps as the regression targets and enforce $l_\text{2}$ loss on all views before and after feature fusion. We train the network for $30$ epochs. Other hyper-parameters such as learning rate and decay strategy are kept the same as in \cite{simplebaselines}. Using a more recent network structure \cite{sun2019deep} generates better $2$D poses.

\begin{figure}
	\centering
	\includegraphics[width=2.6in]{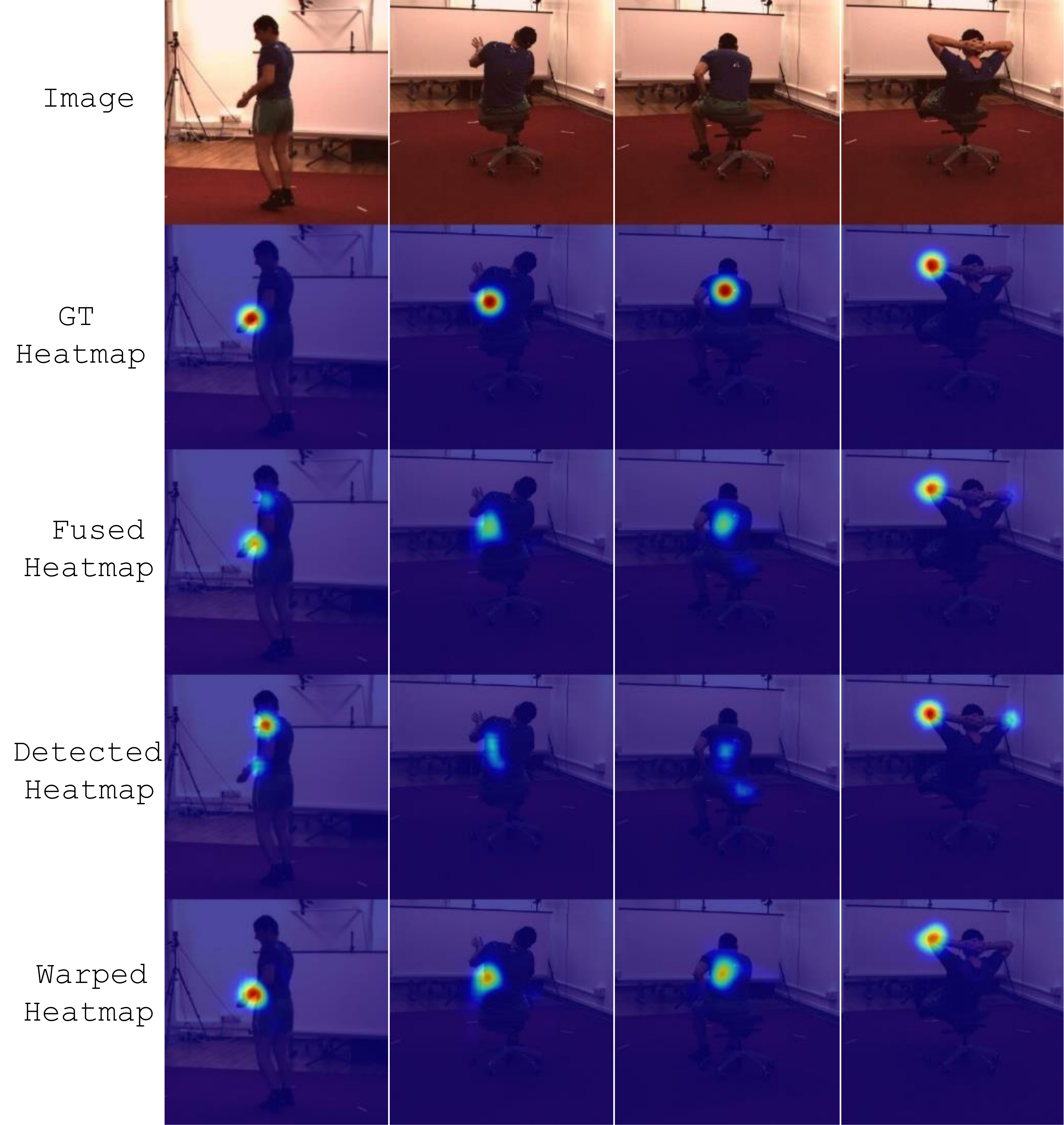}
	\caption{Sample heatmaps of our approach. ``Detected heatmap" denotes it is extracted from the image of the \emph{current} view. ``Warped heatmap" is obtained by summing the heatmaps warped from the other three views. We fuse the ``warped heatmap" and the ``detected heatmap" to obtain the ``fused heatmap". For challenging images, the ``detected heatmaps" may be incorrect. But the ``warped heatmaps" from other (easier) views are mostly correct. Fusing the multi-view heatmaps improves the heatmap quality. }
	\label{fig:heatmap}
\end{figure}

\subsection{Quantitative Results}
Table \ref{table:pose$2$D} shows the results on the most important joints when we train, either only on the H36M dataset, or on a combination of the H36M and MPII datasets. It compares our approach with the baseline method \cite{simplebaselines}, termed \textit{Single}, which does not perform cross view feature fusion. We also compare with two baselines which compute sum or max values over the epipolar line using the camera parameters. The hyper parameters for training the two methods are kept the same for fair comparison.

Our approach outperforms the baseline \textit{Single} on all body joints. The improvement is most significant for the wrist joint, from $85.72\%$ to $95.01\%$, and from $89.33\%$ to $97.20\%$, when the model is trained only on ``H36M" or on ``H36M + MPII'', respectively. We believe this is because ``wrist'' is the most frequently occluded joint and cross view fusion fuses the features of other (visible) views to help detect them. See the third column of Figure \ref{fig:heatmap} for an example. The right wrist joint is occluded in the current view. So the detected heatmap has poor quality. But fusing the features with those of other views generates a better heatmap. In addition, our approach outperforms the \textit{sum} and \textit{max} baselines. This is because the heatmaps are often noisy especially when occlusion occurs. Our method trains a fusion network to handle noisy heatmaps so it is more robust than getting sum/max values along epipolar lines.

It is also interesting to see that when we only use the H36M dataset for training, the \textit{Single} baseline achieves very poor performance. We believe this is because the limited appearance variation in the training set affects the generalization power of the learned model. However, our fusion approach suffers less from the lack of training data. This is probably because the fusion approach requires the features extracted from different views to be consistent following a geometric transformation, which is a strong prior to reduce the risk of over-fitting to the training datasets with limited appearance variation. 

The improved $2$D pose estimations in turn help significantly reduce the error in $3$D. We estimate $3$D poses by direct triangulation in this experiment. Table \ref{table:pose$3$D} shows the $3$D estimation errors on the six important joints.  The error for the wrist joint (which gets the largest improvement in $2$D estimation) decreases significantly from $64.18$mm to $34.28$mm. The improvement on the ankle joint is also as large as $15$mm. The \textit{mean} per joint position error over \textit{all} joints (see (c) and (g) in Table \ref{table:baselines3d}) decreases from $36.28$mm to $27.90$mm when we do not align the estimated $3$D pose to the ground truth.

\subsection{Qualitative Results}
In addition to the above numerical results, we also qualitatively investigate in what circumstance our approach will improve the $2$D pose estimations over the baseline. Figure \ref{fig:heatmap} shows four examples. First, in the fourth example (column), the detected heatmap shows strong responses at both left and right elbows because it is hard to differentiate them for this image. From the ground truth heatmap (the second row) we can see that the left elbow is the target. The heatmap warped from other views (fifth row) correctly localizes the left joint. Fusing the two heatmaps gives better localization accuracy. Second, the third column of Figure \ref{fig:heatmap} shows the heatmap of the right wrist joint. Because the joint is occluded by the human body, the detected heatmap is not correct. But the heatmaps warped from the other three views are correct because it is not occluded there.

\begin{table*}[]
\center
\caption{$3$D pose estimation errors MPJPE ($mm$) of different methods on the H36M dataset. The naming convention of the methods follows the rule of ``A-B-C'' where ``A'' indicates whether we use fusion in $2$D pose estimation. ``Single'' means the cross view fusion is not used. ``B'' denotes the training datasets. ``H36M'' means we only use the H36M dataset and ``+MPII'' means we combine H36M with MPII for training. ``C'' represents the method for estimating $3$D poses.}
\label{table:baselines3d}
\begin{tabular}{l c c c c c c c c}
\toprule
& Direction & Discus & Eating & Greet & Phone & Photo & Posing & Purch \\
\toprule
(a) Single-H36M-Triangulate & 71.76 & 65.89 & 56.63 & 136.52 & 59.32 & 96.30 & 46.67 & 110.51\\
(b) Single-H36M-RPSM & 33.38 & 36.36 & 27.13 & 31.14 & 31.06 & 30.28 & 28.59 & 41.03\\
(c) Single-``+MPII"-Triangulate & 33.99 & 32.87 & 25.80 & 29.02 & 34.63 & 26.64 & 28.42 & 42.63\\
(d) Single-``+MPII"-RPSM & 26.89 & 28.05 & 23.13 & 25.75 & 26.07 & 23.45 & 24.41 & 34.02\\
(e) Fusion-H36M-Triangulate & 34.84 & 35.78 & 32.70 & 33.49 & 34.44 & 38.19 & 29.66 & 60.72\\ 
(f) Fusion-H36M-RPSM & 28.89 & 32.46 & 26.58 & 28.14 & 28.31 & 29.34 & 28.00 & 36.77\\ 
(g) Fusion-``+MPII"-Triangulate & 25.15 & 27.85 & 24.25 & 25.45 & 26.16 & 23.70 & 25.68 & 29.66\\
(h) Fusion-``+MPII"-RPSM & 23.98 & 26.71 & 23.19 & 24.30 & 24.77 & 22.82 & 24.12 & 28.62\\

\toprule
& Sitting & SittingD & Smoke & Wait & WalkD & Walking & WalkT & Average \\
\toprule
(a) Single-H36M-Triangulate  & 150.10 & 57.01 & 73.15 & 292.78 & 49.00 & 48.67 & 62.62 & 94.54\\
(b) Single-H36M-RPSM  & 245.52 & 33.74 & 37.10 & 35.97 & 29.92 & 35.23 & 30.55 & 47.82\\
(c) Single-``+MPII"-Triangulate & 88.69 & 36.38 & 35.48 & 31.98 & 27.43 & 32.42 & 27.53 & 36.28\\ 
(d) Single-``+MPII"-RPSM & 39.63 & 29.26 & 29.49 & 27.25 & 25.07 & 27.82 & 24.85 & 27.99\\
(e) Fusion-H36M-Triangulate & 53.10 & 35.18 & 40.97 & 41.57 & 31.86 & 31.38 & 34.58 & 38.29\\ 
(f) Fusion-H36M-RPSM & 41.98 & 30.54 & 35.59 & 30.03 & 28.33 & 30.01 & 30.46 & 31.17\\ 
(g) Fusion-``+MPII"-Triangulate & 40.47 & 28.60 & 32.77 & 26.83 & 26.00 & 28.56 & 25.01& 27.90\\
(h) Fusion-``+MPII"-RPSM  & 32.12 & 26.87 & 30.98 & 25.56 & 25.02 & 28.07 & 24.37 & 26.21 \\
\toprule
\end{tabular}

\end{table*}

\section{Experiments on $3$D Pose Estimation}

\begin{table}[]
\center
\small
\caption{$3$D pose estimation errors when different numbers of iterations $t$ are used in RPSM. When $t=0$, RPSM is equivalent to PSM. ``+MPII" means we use the combined dataset ``H36M+MPII" to train the 2D pose estimation model. The MPJPE ($mm$) are computed when no rigid alignment is performed between the estimated pose and the ground truth. }
\label{table:rps}
{\scriptsize{
\begin{tabular}{l|c|c|c|c|c}
\hline
Methods & ${t=0}^{\star}$ & $t=1$ & $t=3$ & $t=5$ & $t=10$ \\
\hline
\hline
Single-H36M-RPSM & 95.23 & 77.95 & 51.78 & 47.93 & 47.82 \\
Single-``+MPII"-RPSM & 78.67 & 58.94 & 32.39 & 28.04 & 27.99 \\
Fusion-H36M-RPSM & 80.77 & 61.11 & 35.75 & 31.25 & 31.17 \\
Fusion-``+MPII"-RPSM & 77.28 & 57.22 & 30.76 & 26.26 & 26.21 \\
\hline
\end{tabular}

}}
\end{table}

\subsection{Implementation Details}
In the first iteration of RPSM ($t=0$), we divide the space of size $2,000$mm around the estimated location of the root joint  into $16^3$ bins, and estimate a coarse $3$D pose by solving Eq. \ref{eq:psm}. We also tried to use a larger number of bins, but the computation time becomes intractable.

For the following iterations where $t \ge 1$, we divide the space, which is of size $s_t=\frac{2000}{16 \times 2^{(t-1)}}$, around \textit{each estimated joint location} into $2 \times 2 \times 2$ bins. Note that the space size $s_t$ of each joint equals to the size of a single bin in the previous iteration. We use a smaller number of bins here than that of the first iteration, because it can significantly reduce the time for on-the-fly computation of the pairwise potentials. In our experiments, repeating the above process for ten iterations only takes about $0.4$ seconds. This is very light weight compared to the first iteration which takes about $8$ seconds.

\subsection{Quantitative Results}
We design eight configurations to investigate different factors of our approach. Table \ref{table:baselines3d} shows how different factors of our approach decreases the error from $94.54$mm to $26.21$mm.

\textbf{RPSM vs. Triangulation:}
 First, RPSM achieves significantly smaller 3D errors than Triangulation when $2$D pose estimations are obtained by a relatively weak model. For instance, by comparing the methods (a) and (b) in Table \ref{table:baselines3d}, we can see that, given the same $2$D poses, RPSM significantly decreases the error, \ie from $94.54$mm to $47.82$mm. This is attributed to the joint optimization of all nodes and the recursive pose refinement. 

Second, RPSM provides marginal improvement when $2$D pose estimations are already very accurate. For example, by comparing  the methods (g) and (h) in Table \ref{table:baselines3d} where the $2$D poses are estimated by our model trained on the combined dataset (``H36M+MPII"), we can see the error decreases slightly from $27.90$mm to $26.21$mm. This is because the input $2$D poses are already very accurate and direct triangulation gives reasonably good $3$D estimations. But if we focus on some difficult actions such as ``sitting'', which gets the largest error among all actions, the improvement resulted from our RPSM approach is still very significant (from $40.47$mm to $32.12$mm). 

In summary, compared to triangulation, RPSM obtains comparable results when the $2$D poses are accurate, and significantly better results when the 2D poses are inaccurate which is often the case in practice.

\textbf{RPSM vs. PSM: }
We investigate the effect of the recursive $3$D pose refinement. Table \ref{table:rps} shows the results. First, the poses estimated by PSM, \ie RPSM with $t=0$, have large errors resulted from coarse space discretization. Second, RPSM consistently decreases the error as $t$ grows and eventually converges.
For instance, in the first row of Table \ref{table:rps}, RPSM decreases the error of PSM from $95.23$mm to $47.82$mm which validates the effectiveness of the recursive 3D pose refinement of RPSM.

\textbf{Single vs. Fusion: }
We now investigate the effect of the cross-view feature fusion on 3D pose estimation accuracy. Table \ref{table:baselines3d} shows the results. First, when we use H36M+MPII datasets (termed as ``+MPII") for training and use triangulation to estimate $3$D poses, the average $3$D pose error of our fusion model (g) is smaller than the baseline without fusion (c). The improvement is most significant for the most challenging ``Sitting'' action whose error decreases from $88.69$mm to $40.47$mm. The improvement should be attributed to the better $2$D poses resulted from cross-view feature fusion. We observe consistent improvement for other different setups. For example, compare the methods (a) and (e), or the methods (b) and (f).

\textbf{Comparison to the State-of-the-arts:}
We also compare our approach to the state-of-the-art methods for multi-view human pose estimation in Table \ref{table:state_of_art_3D}. Our approach outperforms the state-of-the-arts by a large margin. First, when we train our approach only on the H36M dataset, the MPJPE error is $31.17$mm which is already much smaller than the previous state-of-the-art \cite{tome2018rethinking} whose error is $52.80$mm. As discussed in the above sections, the improvement should be attributed to the more accurate $2$D poses and the recursive refinement of the $3$D poses.

 \begin{table}
\center
\small
\caption{Comparison of the $3$D pose estimation errors MPJPE ($mm$) of the state of the art multiple view pose estimators on the H36M datasets. We do NOT use the Procrustes algorithm to align the estimations to the ground truth. The result of ``Multi-view Martinez" is reported in \cite{tome2018rethinking}. The four state-of-the-arts do not use MPII dataset for training. So they are directly comparable to our result of $31.17$mm. }
\label{table:state_of_art_3D}
\begin{tabular}{c c}
\toprule
Methods & Average MPJPE \\
\toprule
PVH-TSP \cite{trumble2017total} & $87.3$mm\\
Multi-View Martinez \cite{martinez2017simple} & $57.0$mm\\
Pavlakos \etal \cite{PavlakosZDD17} & $56.9$mm\\
Tome \etal \cite{tome2018rethinking} & $52.8$mm\\
\textbf{Our approach} & \textbf{31.17}mm\\
\textbf{Our approach + MPII} & \textbf{26.21}mm\\
\toprule
\end{tabular}
\end{table}

\begin{figure}
	\centering
	\includegraphics[width=3in]{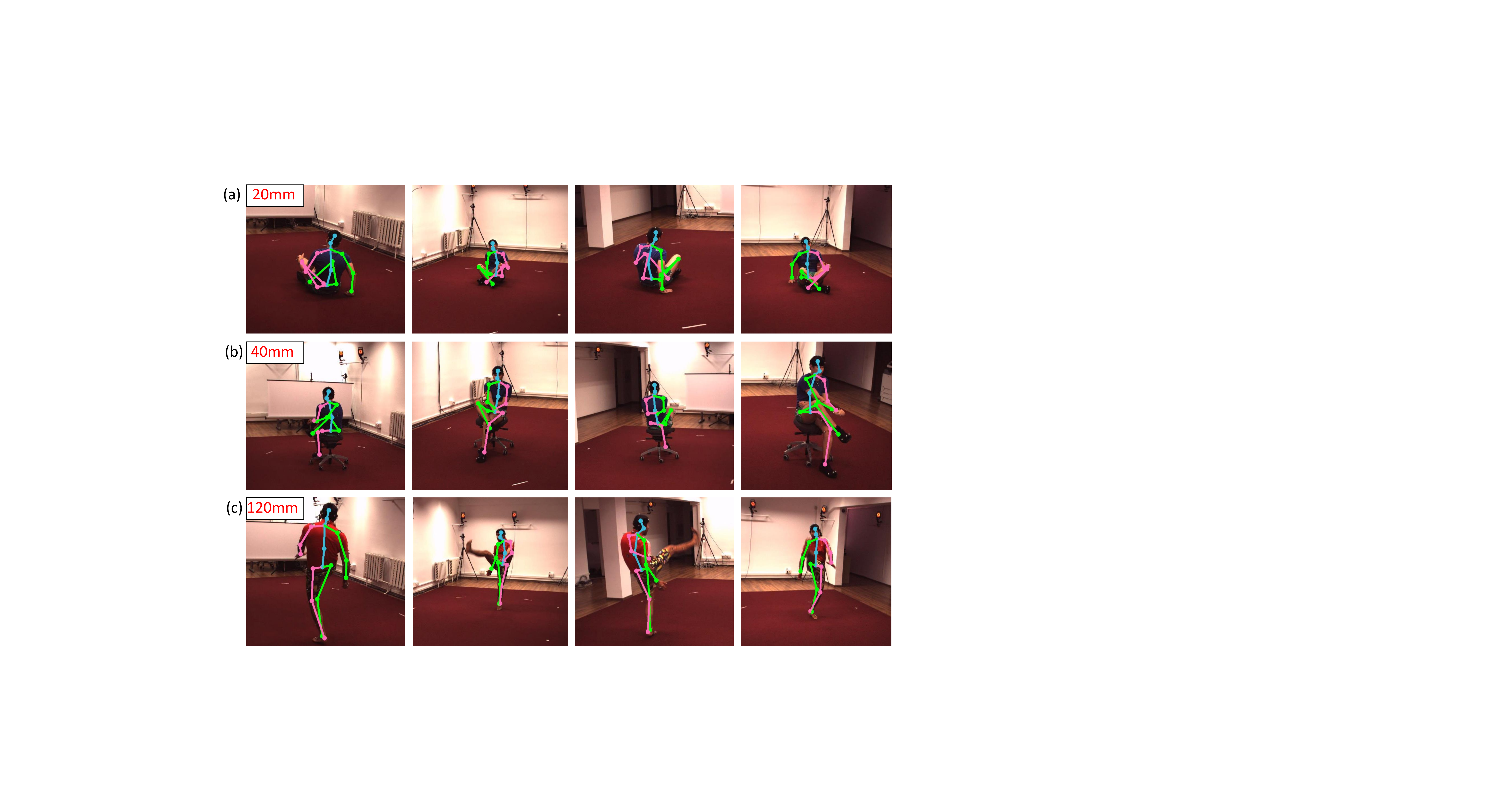}
	\caption{We project the estimated $3$D poses back to the $2$D image space and draw the skeletons on the images. Each row shows the skeletons of four camera views. We select three typical examples whose $3$D MPJPE errors are $20, 40, 120$mm, respectively.}
	\label{fig:visual_$3$D}
\end{figure}

\subsection{Qualitative Results}
Since it is difficult to demonstrate a $3$D pose from all possible view points, we propose to visualize it by projecting it back to the four camera views using the camera parameters and draw the skeletons on the images.
Figure \ref{fig:visual_$3$D} shows three estimation examples.
\emph{According to the 3D geometry, if the 2D projections of a 3D joint are accurate for more than two views (including two), the 3D joint estimation is accurate.} For instance, in the first example (first row of Figure \ref{fig:visual_$3$D}), the 2D locations of the right hand joint in the first and fourth camera view are accurate. Based on this, we can infer with high confidence that the estimated 3D location of the right hand joint is accurate.

 In the first example (row), although the right hand joint is occluded by the human body in the second view (column), our approach still accurately recovers its $3$D location due to the cross view feature fusion. Actually, most leg joints are also occluded in the first and third views but the corresponding 3D joints are estimated correctly. 
 
 The second example gets a larger error of $40$mm because the left hand joint is not accurately detected. This is because the joint is occluded in too many (three) views but only visible in a single view. Cross-view feature fusion contributes little in this case. For most of the testing images, the $3$D MPJPE errors are between $20$mm to $40$mm. 
 
 There are few cases (about $0.05\%$) where the error is as large as $120$mm. This is usually when ``double counting" happens. We visualize one such example in the last row of Figure \ref{fig:visual_$3$D}.  Because this particular pose of the right leg was rarely seen during training, the detections of the right leg joints fall on the left leg regions consistently for all views. In this case, the warped heatmaps corresponding to the right leg joints will also fall on the left leg regions thus cannot drag the right leg joints to the correct positions.

\subsection{Generalization to the Total Capture Dataset}
We conduct experiments on the Total Capture dataset to validate the general applicability of our approach. Our model is trained only on the Total Capture dataset. Table \ref{table:totalcapture} shows the results. ``Single-RPSM" means we do NOT perform cross-view feature fusion and use RPSM for recovering 3D poses. 
 First, our approach decreases the error of the previous best model \cite{trumble2018deep} by about $17\%$. Second, the improvement is larger for the hard cases such as ``FS3". The results are consistent with those on the H36M dataset. Third, by comparing the approaches of ``Single-RPSM'' and ``Fusion-RPSM'', we can see that fusing the features of different views improves the final $3$D estimation accuracy significantly. In particular, the improvement is consistent for all different subsets.

\begin{table}[]
\center
\small
\caption{$3$D pose estimation errors MPJPE ($mm$) of different methods on the Total Capture dataset. The numbers reported for our method and the baselines are obtained without rigid alignment.}
\label{table:totalcapture}
\scriptsize{
\begin{tabular}{l c c c  c c c c}
\toprule
Methods & \multicolumn{3}{c}{Subjects1,2,3} & \multicolumn{3}{c}{Subjects4,5} & Mean\\
& W2 & FS3 & A3 & W2 & FS3 & A3 &  \\
\hline
Tri-CPM \cite{wei2016convolutional} & 79 & 112 & 106 & 79 & 149 & 73 & 99\\
PVH \cite{trumble2017total} & 48 & 122 & 94 & 84 & 168 & 154 & 107 \\
IMUPVH \cite{trumble2017total} & 30 & 91 & 49 & 36 & 112 & 10 & 70\\
AutoEnc \cite{trumble2018deep} & 13 & 49 & 24 & 22 & 71 & 40 & 35 \\
\hline
Single-RPSM & 28 &	42 & 30 & 45 & 74 & 46 & 41\\
\textbf{Fusion-RPSM} & 19 &	28 & 21 & 32 & 54 & 33 & \textbf{29}\\
\toprule
\end{tabular}
}
\end{table}

\subsection{Generalization to New Camera Setups}
We conduct experiments on the H36M dataset using NO pose annotations. The single view pose estimator \cite{simplebaselines} is trained on the MPII dataset. If we directly apply this model to the test set of H36M and estimate the $3$D pose by RPSM, the MPJPE error is about $109$mm. If we retrain this model (without the fusion layer) using the harvested pseudo labels, the error decreases to $61$mm. If we train our fusion model with the pseudo labels described above, the error decreases to $43$mm which is already smaller than the previous supervised state-of-the-arts. The experimental results validate the feasibility of applying our model to new environments without any manual label. 

\section{Conclusion}
We propose an approach to estimate $3$D human poses from multiple calibrated cameras. The first contribution is a CNN based multi-view feature fusion approach which significantly improves the $2$D pose estimation accuracy. The second contribution is a recursive pictorial structure model to estimate $3$D poses from the multi-view $2$D poses. It improves over the PSM by a large margin. The two contributions are independent and each can be combined with the existing methods.

{\small
\bibliographystyle{ieee_fullname}
\bibliography{egbib}
}

\end{document}